\documentclass[10pt,twocolumn,letterpaper]{article}

\usepackage{cvpr}              
\usepackage{booktabs}  
\usepackage{stfloats}
\usepackage{placeins}
\usepackage[accsupp]{axessibility}
\usepackage{multirow}
\usepackage{comment}
\usepackage{amssymb}
\usepackage{pifont}
\definecolor{cvprblue}{rgb}{0.21,0.49,0.74}
\usepackage[pagebackref,breaklinks,colorlinks,allcolors=cvprblue]{hyperref}

\def\paperID{14} 
\def\confName{CVPR}
\def\confYear{2025}

\title{Stochastic-based Patch Filtering for Few-Shot Learning}

\author{Javier Ródenas\\
AIBA, Departament de \\
 Matemàtiques \&  
Informàtica  \\
Universitat de Barcelona\\
Barcelona, Spain\\
{\tt\small jrodencu33@alumnes.ub.edu}
\and
Eduardo Aguilar\\
Departamento de Ingeniería de \\
Sistemas y Computación \\
Universidad Católica del Norte\\
Antofagasta, Chile\\
AIBA, Universitat de Barcelona\\
Barcelona, Spain\\
{\tt\small eduardo.aguilar@ub.edu}
\and
Petia Radeva\\
 AIBA, Departament de \\
 Matemàtiques \&  
Informàtica  
\\ Institute of Neuroscience, 
\\
Universitat de Barcelona\\
Barcelona, Spain\\
{\tt\small petia.ivanova@ub.edu}
}

\newcommand{\museNo}{01070\-421}
\newcommand{\DFVolNo}{PDC\-2022-133642-I00}

\begin{document}
\maketitle

\begin{abstract}

Food images present unique challenges for few-shot learning models due to their visual complexity and variability. For instance, a pasta dish might appear with various garnishes on different plates and in diverse lighting conditions and camera perspectives. This problem leads to losing focus on the most important elements when comparing the query with support images, resulting in misclassification. To address this issue, we propose Stochastic-based Patch Filtering for Few-Shot Learning (SPFF) to attend to the patch embeddings that show greater correlation with the class representation. The key concept of SPFF involves the stochastic filtering of patch embeddings, where patches less similar to the class-aware embedding are more likely to be discarded. With patch embedding filtered according to the probability of appearance, we use a similarity matrix that quantifies the relationship between the query image and its respective support images. Through a qualitative analysis, we demonstrate that SPFF effectively focuses on patches where class-specific food features are most prominent while successfully filtering out non-relevant patches. We validate our approach through extensive experiments on few-shot classification benchmarks: Food-101, VireoFood-172 and UECFood-256, outperforming the existing SoA methods. 

\end{abstract}

\section{Introduction}

Over the last few years, scaling deep learning models has been a priority for the research community. Adding new food categories to deep learning models without requiring large amounts of data has been constantly being developed. Previously, a large quantity of data was needed, requiring significant resources both in terms of time and computation. To solve this problem,  Few-Shot Learning (FSL) approaches claim to classify a new category with just a few examples \cite{RER, MVFSL-TC}.

Few-shot learning represents an innovative strategy to address the challenge of the limited labeled data. This approach focuses on developing algorithms capable of effectively adapting and generalizing to new categories using a minimal number of training examples, typically Zero-Shot Learning (ZSL) \cite{mirza2023lafter,Stojni?_2024_CVPR}, One-Shot Learning (OSL) \cite{yu2024learningshot, Chen_2019_CVPR, vinyals2017matchingnetworksshotlearning} and Few-Shot Learning (FSL), where we can find shots between one and five per class generally. Research in this field has generated several promising methodologies, including Siamese network architectures \cite{siamese}, memory-based systems \cite{memory}, models based on similarity metrics \cite{semantic} or MALM \cite{MAML} which introduced a parameter initialization to rapidly adapt the model to new tasks with minimal steps.
This paradigm of learning from limited examples facilitates rapid adaptation to novel tasks and enables efficient knowledge acquisition in domains where data collection necessitates substantial resources and specialized expertise. 


However, food image classification presents challenges that differentiate it from other image classification tasks. Food exhibits high intra-class variability due to differences in presentation, illumination, and capture angles. In addition, there is considerable inter-class overlap between similar categories. Diversity in food adds another layer of complexity, as similar dishes can vary signification between regions. These factors make traditional deep learning approaches which require large amounts of labeled data, difficult to implement effectively in this domain. 

A significant limitation of FSL is its inherent susceptibility to overfitting. The scarcity of training data often leads models to memorize the limited training examples rather than extracting generalizable features, a particular problem when dealing with the complexity of food images. The work in \cite{overfitting} presented a comprehensive analysis of overfitting patterns in FSL scenarios and proposed regularization techniques specifically designed to address the challenges of overfitting. The authors of \cite{overfitting2} demonstrate neural networks naturally favor discriminative over transferable features that intensify overfitting in FSL and propose a dual margin architecture setting positive and negative margins to avoid overfitting.

Another major challenge is the problem of noisy features. Conventionally, in machine learning scenarios involving extensive datasets, models naturally learn to discard irrelevant information during the training phase. However, FSL needs attention to discriminating between relevant and non-relevant information due to scarcity of data. This need is particularly significant in tasks such as food image classification, where non-relevant components like background or other elements often cause classification errors. Recent advances in addressing this limitation have emerged in different works. Metric-based approaches, such as \cite{lai2024clusteredpatchelementconnectionfewshot}, implemented sophisticated cosine similarity architectures that selectively filter relevant features from query-support image pairs, removing noise and irrelevant information. In addition, attention approaches, as shown in \cite{wang2023focus}, employ attention and gradient information to automatically locate the positions of key entities in the support images.

\begin{figure*}[t!]
\centering
\includegraphics[width=1\textwidth, height=0.3\textheight]{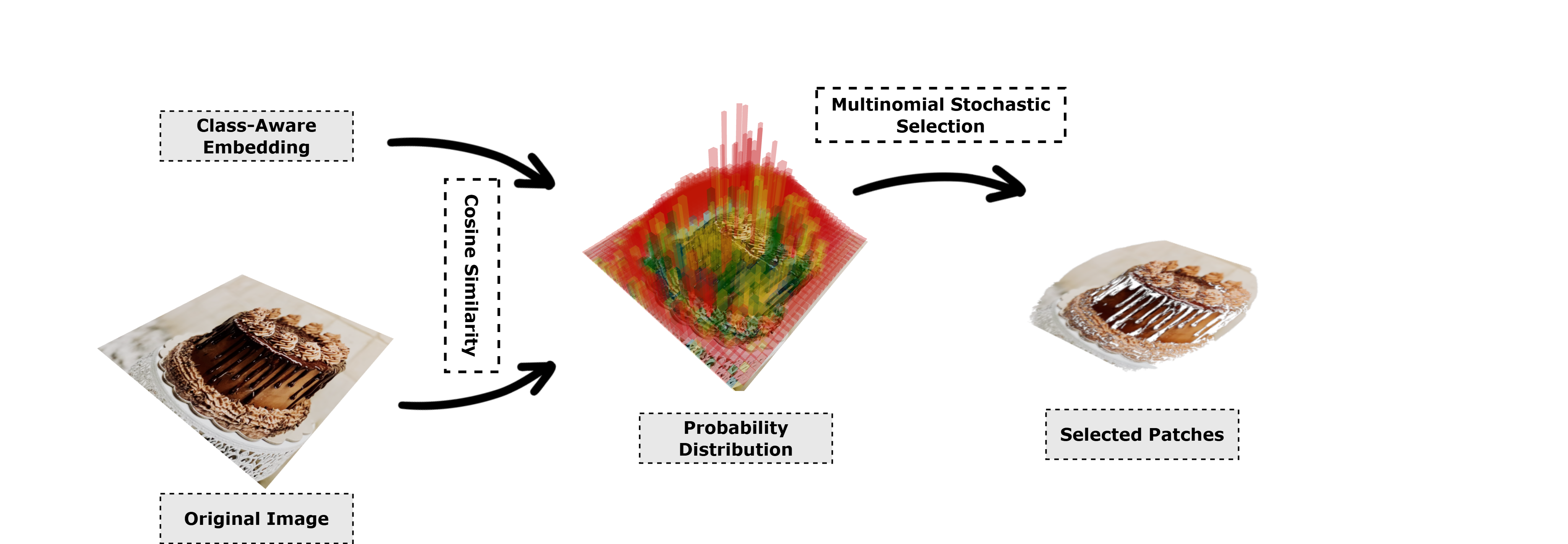}
\caption{Pipeline of SPFF. Starting with an original image, we extract the patch embedding and a class-aware embedding and compute their cosine similarity. This generates a probability distribution representing patch importance. The multinomial stochastic selection then samples patches according to these probabilities.}
\label{fig:Multinomial}
\end{figure*}

Recent research has explored few-shot learning approaches, especially focused on food recognition, addressing the complexities in food images. For example, LR \cite{LR} utilizes graph convolutional networks and semantic embeddings to capture relationships between food categories. More recently, RER \cite{RER} employs an adversarial erasing strategy to automatically discover ingredients in images and learn more comprehensive representations. MVFSL-TC \cite{MVFSL-TC} introduced a multi-view learning model to extract features oriented not just to categories, but to ingredients. 

Unlike these methods that rely on additional information such as ingredients or extra semantic representations, our SPFF approach fundamentally addresses the challenge of distinguishing relevant features from non-relevant ones through stochastic patch selection. Where previous approaches attempted to learn better global representations or optimize similarity metrics, SPFF focuses on modeling the importance of patch embeddings based on class-specific features, enabling our model to focus primarily on the food while filtering out irrelevant background elements. Thus, the final feature representation is enriched with information that is most relevant to the target class. This approach to feature selection is particularly well-suited to food recognition, where class-specific details may occupy only a portion of the image having numerous distractors.

Our method introduces a novel way to address the problem of feature relevance, namely Stochastic-based Patch Filtering for Few-Shot Learning (SPFF), which filters the irrelevant features. Using a class-aware embedding that contains the global class features, SPFF computes the similarity between class-aware embedding and patch embeddings. So, it can assign a probability based on their similarity while leaving a certain degree of stochastic freedom. The stochastic patch selection serves as a filtering strategy to extract patches of the highest relevance as shown in Fig. \ref{fig:Multinomial}. This method guarantees that the final feature representation captures the most critical information related to the target class. Additionally, it enables us to discard irrelevant background noise, emphasizing the distinctive features that characterize each category. By selecting an optimal number of patches, we retain enough information to avoid losing essential details while effectively filtering out less relevant patches. Finally, SPFF adopts a fusion strategy of linear addition, which allows one to integrate class-aware representation. This approach effectively alleviates the scarcity of labeled images by enriching the feature set with additional class-relevant patches. We validate our method to three public datasets, namely: Food-101 \cite{food101}, VireoFood-172 \cite{VireoFood172} and UECFood-256 \cite{UECFood256} with very encouraging results.

Our contributions are summarized as follows:

\begin{itemize}
 \item We propose SPFF, a novel method that selects stochastically patches with the most relevant information while filtering irrelevant ones.
 \item We demonstrate the importance of stochastic selection over deterministic approaches, showing that controlled randomness helps capture more diverse and discriminative patches.
 \item We validate SPFF extensively on Food-101, VireoFood-172, and UECFood-256 datasets, demonstrating improved performance over current state-of-the-art methods.
 \item We provide comprehensive visual analyses, clearly illustrating that SPFF successfully isolates relevant class-specific features while discarding non-relevant patches.
\end{itemize}

The paper is structured as follows. In Section \ref{sec:related_work}, we present the related work to our method. Section \ref{sec:methodology} defines our method SPFF and its different steps. In Section \ref{sec:experiments}, we show the results on three food public datasets and comparison to the state-of-the-art. Finally, conclusions are given in Section \ref{sec:conclusions}, where we share the final thoughts, limitations and future works.

\section{Related Work}
\label{sec:related_work}

In this section, we briefly discuss the most recent and relevant work on FSL and feature filtering techniques.

\subsection{Few-shot Learning}

In recent years, FSL has emerged as one of the most critical areas of research in computer vision. Existing methods can be categorized into metric-based, optimization-based methods, and transfer learning-based methods.

In the \textbf{optimization-based methods}, the models focus on finding ways to optimize the learning process and adapt to generalize properly to new tasks. MAML \cite{MAML} is one of the most popular methods that focuses on learning a set of optimal parameters given a small number of samples from a new task. A similar method, Reptile \cite{Reptile}, simplifies the meta-learning process by reducing the computational cost of updating the optimal parameters for every task.

The main idea of \textbf{metric-based FSL methods} is to obtain the most representative embedding from query and support images to ensure that the samples of the same class are close to each other. Then, they are compared by calculating the distance using a similarity metric. First, they focus on learning the feature space to get the best feature representation. So that, the similarity metric is used to distinguish the most similar support for a given query \cite{NIPS2017_cb8da676, Hao_2023_ICCV, vinyals2017matchingnetworksshotlearning}.

 \textbf{Transfer learning based methods} are trained on a large dataset to learn general features of a related domain. These models are used as feature extractors to get transferable representations and fine-tune a new task. This approach allows one to learn faster, with less data and resulting in a better performance due to the transferred knowledge. For example, MTL \cite{MTL} leverages a large number of similar few-shot tasks in order to learn how to adapt a pre-trained model to a new task with only a few labels.

\subsection{Feature Filtering}
Due to the scarcity of data in FSL problems, feature filtering is very important to learn the best representation using the most relevant features. However, feature filtering presents several challenges. First, the feature importance patterns learned from one task may not generalize across multiple tasks. Second, the scarcity of examples makes the relevant features identification very challenging \cite{aligment, taskguided}. So, the combination of scarce data and potentially irrelevant features increases the risk of overfitting or misclassification due to features as background \cite{graph, semantic}. 

Recent advances in feature filtering strategies have emerged to address these challenges. \cite{Hilbert} introduced a method for few-shot learning using the Hilbert-Schmidt Independence Criterion to select relevant features in FSL scenarios. Furthermore, attention mechanisms have shown remarkable effectiveness in enhancing feature representation quality, particularly scenarios with limited data \cite{dual, rectification, SaberNet}. These mechanisms enhance the model's ability to focus on both global and localized features, helping to filter out irrelevant information and ensuring that the representation learned is robust and task-specific.

Lately, Vision Transformers (ViT) have emerged as prominent architectural paradigms within FSL contexts and different methods employ a pre-trained ViT model. FewTURE \cite{FewTURE} showed that a purely ViT-based architecture can be effectively adapted to limited-scale datasets through a methodical partitioning of inputs by dividing them into patches. Also, CPEA \cite{Hao_2023_ICCV} leverages a pre-trained ViT model through the direct integration of patch-level embeddings with class-aware semantic representations. CPEA employs a homogeneous weighting mechanism when computing similarities between visual patch embeddings and class-specific representational vectors. More recently, CPES \cite{select} introduces a novel approach for selecting class-relevant patch embeddings to address the fundamental challenges of background inference in few-shot classification. This method filters patch embeddings using similarity with class embeddings to retain only the most relevant ones.

\section{Methodology} \label{sec:methodology}
In this section, we first define the problem of few-shot learning and present the whole pipeline. Then, we explain in detail the feature filtering mechanism. Finally, we describe the classification method based on similarities between query and support patch-filtered embeddings.

\subsection{Problem Definition}

The few-shot image classification problem is defined as a \textit{N}-way \textit{M}-shot task, where \textit{N} indicates the number of classes and \textit{M} indicates the number of labeled images per class. This task includes a support set \textit{S} containing \textit{M} labeled images per every \textit{N} class and a query set \textit{Q} containing images to be classified into one of these \textit{N} classes from \textit{S}. 

The support and query set consists of the training set, validation set, and test set, where the test set is fully unseen during the training and validation phases. The goal is to train a model on training classes to generalize well on unseen test classes based on the support set images, using the query set to evaluate the performance on the defined task. In practice, the matching process between support and query images is not straightforward due to the limited number of labeled examples per class \textit{M}-shot meaning that models learn with minimal data, often lacking diversity and variation.

\subsection{SPFF Architecture Overview}

\begin{figure*}[t!]
\centering
\includegraphics[width=0.95\textwidth]{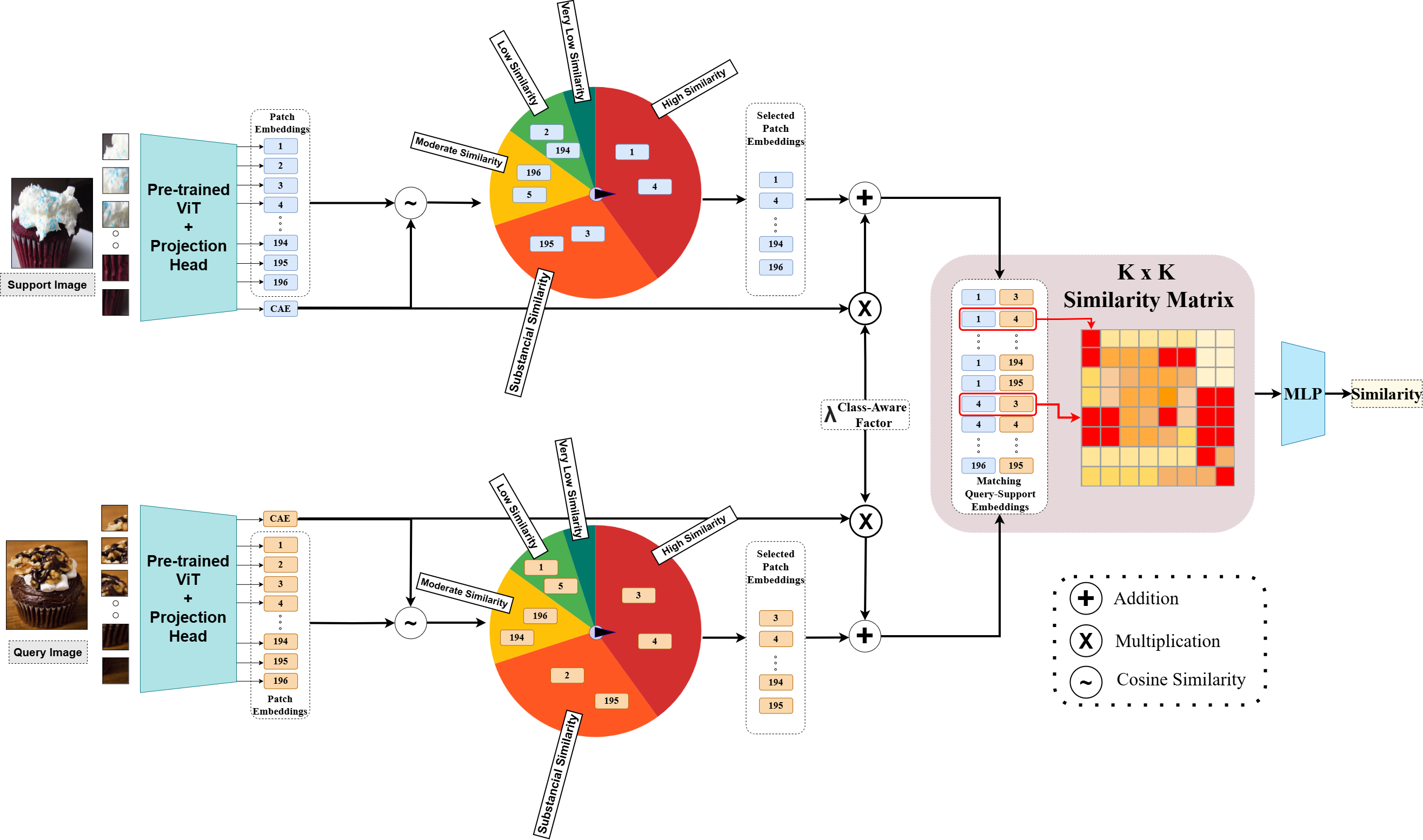}
\caption{Illustration of the SPFF pipeline. First, the ViT-S/16 captures both support and query image information from the input image, generating embeddings for each patch and also feeding the class-agnostic embedding learning (CAE) the class representation. In the second part, SPFF refines the patch embeddings by using stochastic patch selection which is in charge of giving probabilities to the patches based on the similarity with class-aware embedding. With probabilities assigned, the most relevant patches are selected based on a multinomial distribution. Finally, the class-aware embeddings are additionally added with a class-aware factor and the similarity between the query and support refined patches is computed to get the final score.}
\label{fig:pipeline}
\end{figure*}

The pipeline of our method can be seen in Fig.~\ref{fig:pipeline}. Our method works as follows: first, the input images are processed through a pre-trained Vision Transformer (ViT) backbone used as a feature extractor. The ViT splits each image into patches of (16 x 16 pixels), embeds them through linear projection, and gets a class representation capturing global image information. This process generates a sequence of patch $embeddings \in \mathbb{R}^{P\times D}$ where $P=196$ is the number of patches and $D=384$ is the embedding dimension. 

Then, cosine similarity between patch embeddings and class token is computed. The similarity scores are used to assign importance weights to each patch, creating an attention mechanism that highlights the most relevant features. These weights are converted into a probability distribution using a softmax function.

Next, SPFF employs a stochastic selection process in which patches are sampled according to their probability weights using a multinomial distribution. This introduces a controlled randomness that helps the model focus on different patches across training iterations, improving its generalizability.

The selected patches from both support and query images are refined and aligned. The class-aware embeddings are used to enhance patches with an additional class-aware factor that emphasizes class-specific features.

Finally, we compute a dense similarity matrix between the selected patch embeddings of query and support images. This matrix is processed through a Multi-Layer Perceptron (MLP)  that converts the similarities to scores.

\subsection{Stochastic-based Patch Filtering}

After obtaining the patch embeddings from ViT, we implement our Stochastic-based patch filtering mechanism to identify and focus on the most important features of the image. 

Given a set of patch embeddings $P = \{P_1, P_2, P_3..., P_m\} \in \mathbb{R}^{B \times N \times D}$ where $B$ is the batch size, $N=196$ the number of patch embeddings and $D=384$ embedding dimension, we compute their similarity with the class token $C \in \mathbb{R}^{B \times 1 \times D}$. First, we normalize patch embeddings and the class token using L2 normalization:

\begin{equation}
\hat{P} = \frac{P}{||P||_2}, \quad \hat{C} = \frac{C}{||C||_2}.
\end{equation}

Then, we compute the cosine similarity between $\hat{{\text{P}}}$ and $\hat{\text{C}}$ as follows:

\begin{equation} S_i = \hat{P}_i \cdot \hat{C}, \quad \forall i \in {1, 2, ..., P}, \end{equation}

where $S_i$  represents the similarity score for the i-th patch. These similarity scores indicate how closely each patch aligns with the global class representation captured by the class token.

Next, we convert these similarity scores into a probability distribution using a softmax function:

\begin{equation}
p_i = \frac{\exp(S_i)}{\sum_{j=1}^{P} \exp(S_j)}.
\end{equation}

Instead of deterministically selecting the top-k patches based on these probabilities, we introduce stochasticity by sampling patches according to a multinomial distribution parameterized by the calculated probabilities:

\begin{equation}
{i_1, i_2, ..., i_k} \sim \text{Multinomial}(k, p),
\end{equation}

where k is the number of patches to be selected, $p={p1,p2,...,pk}$ is the vector of probabilities and $I_{selected}={i1,i2,...,ik}$ is a vector of indices of selected patch embeddings. Filtering patches according to the indices:

\begin{equation}
P_{selected} = {p_{i_1}, p_{i_2}, ..., p_{i_k}}.
\end{equation}

This patch embedding selection allows the model to explore different patch embeddings across training iterations, prevents overfitting by introducing controlled randomness, and improves the robustness of the model. The selected patch embeddings form a refined representation that focuses on the most relevant features while filtering out non-relevant patch embeddings.

Once the embeddings are filtered, we proceed with the linear class-aware addition:
\begin{equation}
\hat{P} = P_{selected} + \lambda \cdot {C},
\end{equation}

where $\lambda$ is a factor that controls the influence of the class token. That means that we are forcing the patches to have a class-relevant representation. We set $\lambda=2$ following the strategy validated in \cite{Hao_2023_ICCV}.

\subsection{Classification}

After obtaining the filtered features, we compute similarities between query and support samples that are performed through a dense score matrix:
\begin{equation}
T_{ij} = d(\hat{P}{support}, \hat{P}{query})
\end{equation}
where $d(·, ·)$ denotes the cosine similarity. 

Next, we flatten the similarity matrix $T$ processed through an MLP to obtain the final classification scores: 

\begin{equation}
scores_{ij} = MLP(Flatten(T_{ij}))
\end{equation}
where $T_{ij}$ is the similarity matrix between query image $i$ and support image $j$.

For the N-way M-shot classification, we aggregate the similarity scores across the $M$ support samples for each class. For query image $i$ and class $n$, the aggregated score, $s_{i}^{n}$ is:
\begin{equation}
s^n_{i} = \sum_{m=1}^M scores^n_{im}
\end{equation}
where $scores^n_{im}$ represents the similarity between query image $i$ and the $m$-th support image of class $n$.

Finally, we compute the classification probabilities (p) using the softmax function:
\begin{equation}
p^n_{i} = \frac{\exp(s^n_{i})}{\sum_{m=1}^N \exp(s^m_{i})}
\end{equation}

The model is trained by minimizing the cross-entropy loss over all query images.

\section{Experiments}\label{sec:experiments}

In this section, we provide an introduction to the datasets used and the implementation pipeline. We then discuss the performance of our approach, compare it with the state-of-the-art methods, and provide an extensive analysis of the results obtained.

\subsection{Datasets}
To evaluate the effectiveness of our proposed SPFF method, we utilize three publicly available datasets commonly employed in few-shot learning tasks: 1) {\bf Food-101} \cite{food101}, consists of 101,000 images distributed across 101 food categories, with 1,000 images per category; 2) {\bf VireoFood-172} \cite{VireoFood172} contains 172 food categories with a total of 110,241 images; and 3) {\bf UECFood-256} \cite{UECFood256} features 256 food categories with a total of 31,397 images.

Consistent with previous approaches \cite{LR, Hao_2023_ICCV}, we split each dataset into 70\% for training, 10\% for validation, and 20\% for testing. The label spaces are non-overlapping, ensuring that classes present in the training set do not appear in the validation or test sets.

\begin{table*}[!htbp]
\centering
\caption{Comparison with state-of-the-art methods on 5-way 1-shot and 5-way 5-shot with 95\% confidence intervals on Food-101, VireoFood-172 and UECFood-256.}
\label{tab:comparison_sota}
\resizebox{\textwidth}{!}{
\begin{tabular}{lccccccccc}
\toprule
\multirow{2}*{Model} & \multirow{2}*{Backbone} & \multirow{2}*{$\approx$ \# Params} & \multicolumn{2}{c}{Food-101} & \multicolumn{2}{c}{VireoFood-172} & \multicolumn{2}{c}{UECFood-256} \\
\cmidrule(lr){4-5} \cmidrule(lr){6-7} \cmidrule(lr){8-9}
& & & 1-shot & 5-shot & 1-shot & 5-shot & 1-shot & 5-shot\\
\midrule
MVFSL-LC \cite{MVFSL-TC} & VGG-16 & 138 M & 55.10 & 68.10 & 74.80 & 83.50 & - & - \\
\midrule
MVFSL-TC \cite{MVFSL-TC} & VGG-16 & 138 M & 55.30 & 68.30 & 75.10 & 83.60 & - & - \\
\midrule
Fusion Learning \cite{LR} & EfficientNet-B0 & 5.3 M & 61.97 & 77.72 & - & - & 54.85 & 69.08 \\
\midrule
RER \cite{RER} & ResNet-12 & 12.4 M & 62.51 & 80.47 & 82.13 & 93.76 & - & - \\

\midrule
\textbf{SPFF (ours)} & ViT-S/16 & 22 M & \textbf{65.07} & \textbf{83.32} & \textbf{82.31} & \textbf{94.64} & \textbf{73.82} & \textbf{88.71} \\
\bottomrule
\multicolumn{7}{l}{}
\end{tabular}
}
\end{table*}

\begin{figure*}[t!]
\centering
\includegraphics[width=0.52\textwidth, height=0.450\textheight]{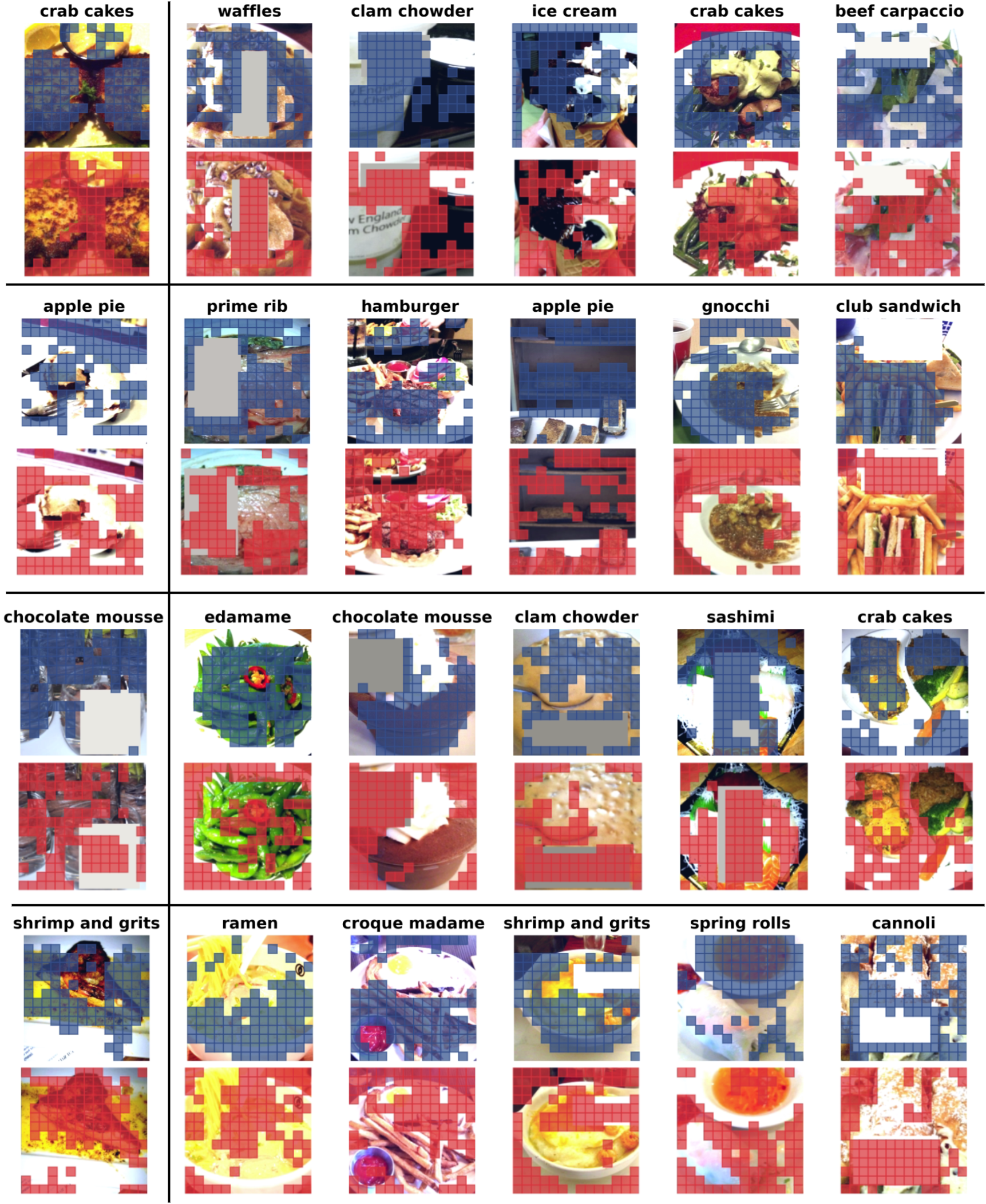}
\caption{Example of visualization comparing the stochastic patch selection (first row, blue overlays) versus the deterministic approach (second row, red overlays) on various Food-101 dishes. The query images appear on the left, and to the right of the vertical dividing line are the support images. The blue highlights in the first row indicate fully stochastic patch selection, while the red highlights in the second row indicate a purely deterministic strategy.}
\label{fig:example_SPFF}
\end{figure*}

\subsection{Experimental Settings}

We use a ViT-B/16 \cite{VIT16} pretrained using the \cite{tokenizer} strategy. We followed the choice of the backbone and hyperparameters settings established by \cite{Hao_2023_ICCV} in order to compare correctly and fairly. The number of selected patches is set to 98, which represents 50\% of the total of 196 patches in the ViT architecture. This selection rate provides an optimal balance between computational efficiency and representational capacity, allowing our model to focus on the most relevant patches while filtering out non-relevant ones. Our experimental evaluation follows the standard few-shot learning protocol \cite{Hao_2023_ICCV, augmentation}, conducting tests in 5-way 1-shot and 5-way 5-shot scenarios. For each dataset, we evaluate the model's performance over 1,000 randomly generated test episodes, where each episode contains 15 query samples per class, and report the mean classification accuracy.

\subsection{Patch Selection Analysis}

We conducted a comprehensive analysis of our patch selection strategy on the Food-101 and UECFood-256 datasets using the 5-way 5-shot setting. These experiments investigate two critical aspects: the balance between deterministic and Stochastic-based patch selection and the impact of varying the total number of selected patches.

\begin{table*}[!htbp]
\centering
\begin{tabular}{cccc}
\toprule
\textbf{Number of Patches (K)} & \textbf{Deterministic} & \textbf{Stochastic} & \textbf{Random} \\
\midrule
32 & 81.44 & 82.34 & -\\
49 & 82.22 & 82.83 & -\\
64 & 82.80 & 83.23 & -\\
98 & \textbf{82.84} & \textbf{83.32} & \textbf{82.05} \\
128 & 82.08 & 82.85 & -\\
164 & 82.11 & 82.54 & -\\
196 & 82.14 & 82.14 & -\\
\bottomrule
\end{tabular}
\caption{Impact of varying the number of selected patches (K) using deterministic, stochastic and random patch selection strategies on Food-101 dataset in the 5-way 5-shot setting.}
\label{tab:patch_number_comparison}
\end{table*}

\begin{table}[th]
\centering
\begin{tabular}{cccc}
\toprule
\multirow{2}{*}{\textbf{Stochastic (\%)}} & \multirow{2}{*}{\textbf{Fix (\%)}} & \multicolumn{2}{c}{\textbf{Accuracy (\%)}} \\
\cmidrule(lr){3-4}
& & \textbf{Food-101} & \textbf{UECFood-256} \\
\midrule
0 & 100 & 82.84 & 87.37 \\
25 & 75 & 82.87 & 87.21 \\
50 & 50 & 82.80 & 88.22 \\
75 & 25 & 83.03 & 87.93 \\
100 & 0 & \textbf{83.32} & \textbf{88.71} \\
\bottomrule
\end{tabular}
\caption{Effect of balancing deterministic (Fix) and stochastic patch selection on SPFF performance with K=98 patches for Food-101 and UECFood-256 datasets in 5-way 5-shot setting.}
\label{tab:fix_random_balance}
\end{table}

\begin{table}[th]
\centering
\begin{tabular}{ccc}
\toprule
\multirow{2}{*}{\textbf{Number of Patches (K)}} & \multicolumn{2}{c}{\textbf{Accuracy (\%)}} \\
\cmidrule(lr){2-3}
& \textbf{Food-101} & \textbf{UECFood-256} \\
\midrule
32 & 82.34 & 87.53 \\
49 & 82.83 & 87.98 \\
64 & 83.23 & 88.48 \\
98 & \textbf{83.32} & \textbf{88.71} \\
128 & 82.85 & 88.65 \\
164 & 82.54 & 88.41 \\
196 & 82.14 & 88.36 \\
\bottomrule
\end{tabular}
\caption{Impact of varying the number of selected patches (K) on SPFF performance with fully stochastic selection for Food-101 and UECFood-256 datasets in 5-way 5-shot setting.}
\label{tab:patch_number}
\end{table}

Table \ref{tab:fix_random_balance} presents the results of our first experiment, where we vary the proportion of patches selected deterministically (based on highest similarity scores) versus stochastically (based on multinomial sampling). The total number of selected patches remains constant at K=98, representing 50\% of the total available patches. Fully stochastic selection achieves the highest accuracy of 83.32\% on Food-101 and 88.71\% on UECFood-256, outperforming all mixed strategies and the fully deterministic approach. This suggests that the controlled randomness introduced by our stochastic selection mechanism effectively prevents overfitting and enhances the generalization capability of the model.

In our second experiment, we investigate the optimal number of patches to select when using fully stochastic selection. Table \ref{tab:patch_number} shows the performance across different values of K. We observe that the accuracy initially increases with K, reaching a peak of 83.32\% on Food-101 and 88.71\%  on UECFood-256 at K=98. This pattern indicates that, while including more relevant patches improves performance up to a certain point, selecting too many patches eventually incorporates noise and redundant information. The performance drop at K=196 further validates the effectiveness of our patch-filtering approach compared to using all patches without filtering, where K=196 means selecting all patches.

\begin{table}[ht]
\centering
\begin{tabular}{lc}
\toprule
\textbf{Metric} & \textbf{Accuracy (\%)} \\
\midrule
Cosine Similarity & \textbf{83.32} \\
Manhattan Distance & 82.90 \\
Euclidean Distance & 82.88 \\
\bottomrule
\end{tabular}
\caption{Comparison of accuracy across different metrics in 5-shot 5-way setting for Food-101.}
\label{tab:distance_comparison}
\end{table}

As illustrated in Figure \ref{fig:example_SPFF}, our visualization highlights how the model selects distinct patches for both the query images (on the left) and the corresponding support images (on the right) using two patch selection strategies. We can infer that the fully stochastic selection (blue) trends cover a broader range of discriminative areas, while the deterministic selection (red) approach often appears more localized and, in some cases, includes patches with less relevance or even noise. For example, in the second row, third column we can see an image labeled as "hamburger" where using stochastic selection highlights not only portions of the burger but also additional context while deterministic selection seems more clustered on the background or table.

To further validate the effectiveness of our stochastic selection approach, we conducted a comparative analysis between deterministic, stochastic and random patch selection methods across various patch counts. As shown in Table \ref{tab:patch_number_comparison}, stochastic selection consistently outperforms deterministic selection for almost all patch counts. The performance gap is particularly pronounced at K=98, where stochastic selection achieves 83.32\% accuracy compared to 82.84\% with deterministic selection, representing 0.48\% improvement. This consistent pattern of improvement confirms that the probabilistic nature of our approach enables the model to focus on more diverse and relevant patches, leading to better generalization. Furthermore, the random approach, where K=96 patches are randomly chosen without taking into account any similarity strategy, yields 82.05\% showing the importance of distance metrics. As shown in Table \ref{tab:distance_comparison}, using cosine similarity achieves the highest accuracy \textbf{83.32\%}.

These analyses provide empirical evidence for two key design choices in our SPFF framework: (1) the superiority of fully stochastic patch selection over deterministic or mixed strategies, and (2) the importance of selecting an optimal number of patches, which balances between capturing sufficient information and avoiding noise.

\subsection{Performance Comparison to the SoTA}

The results demonstrate that SPFF exceeds the state-of-the-art methods as shown in Table \ref{tab:comparison_sota}.  On Food-101, the accuracy reaches \textbf{83.32\%} in the 5-shot setting and \textbf{65.07\%} in the 1-shot setting, improving \textbf{2.85\%} and \textbf{2.56\%} respectively compared to RER \cite{RER}. For UECFood-256, SPFF achieves \textbf{88.71\%} accuracy in the 5-shot setting and \textbf{73.82\%} in the 1-shot setting representing a \textbf{18.97\%} and \textbf{19.63\%}  improvement over LR \cite{LR}. Finally, on VireoFood-172, our model attains \textbf{94.64\%} accuracy for 5-shot tasks, surpassing RER by \textbf{0.88\%} and \textbf{82.31\%} accuracy for 1-shot tasks, surpassing RER by \textbf{0.18\%}. 

These results consistently show that SPFF achieves superior performance across all datasets and settings, with particularly substantial improvements on challenging datasets such as UECFood-256. The consistent gains across different shot settings demonstrate that our stochastic patch filtering mechanism effectively captures discriminative features even with extremely limited examples.

\section{Conclusions}
\label{sec:conclusions}

The  SPFF approach demonstrates superior performance in multiple food benchmark datasets. Through comprehensive evaluation, SPFF achieves significant improvements over state-of-the-art results on Food-101, VireoFood-172, and UECFood-256 datasets, with particularly impressive gains in the challenging UECFood-256 benchmark. 

The experimental results validate the effectiveness of our stochastic-based patch filtering mechanism. The consistent performance improvements across various benchmarks can be attributed to the key aspect of the stochastic patch selection strategy, which effectively enables the model to focus on more diverse and relevant patches leading to better generalization by introducing controlled randomness. 

However, our approach does have limitations. The performance is sensitive to the number of selected patches and its generalization to other domains with different visual characteristics remains to be fully explored. In future work, we plan to extend SPFF to handle more diverse visual domains beyond food datasets and investigate adaptive patch selection strategies that can automatically determine the optimal number of patches based on image content.

\section*{Acknowledgements} This work has been partially supported by the Horizon EU project MUSAE (No. \museNo), 2021-SGR-01094 (AGAUR), Icrea Academia'2022 (Generalitat de Catalunya), Robo STEAM (2022-1-BG01-KA220-VET-000089434, Erasmus+ EU), DeepSense (ACE053/22/000029, ACCIÓ), DeepFoodVol (AEI-MICINN, \DFVolNo), PID2022-141566NB-I00 (AEI-MICINN), and Beatriu de Pinós Programme and the Ministry of Research and Universities of the Government of Catalonia (2022 BP 00257).

\bibliographystyle{ieeenat_fullname}
\bibliography{main}

\end{document}